\documentclass{article}
\PassOptionsToPackage{numbers, compress}{natbib}

\usepackage[preprint]{tackling_climate_workshop_style}

\usepackage[utf8]{inputenc} 
\usepackage[T1]{fontenc}    
\usepackage{hyperref}       
\usepackage{url}            
\usepackage{booktabs}       
\usepackage{amsfonts}       
\usepackage{makecell}       
\usepackage{nicefrac}       
\usepackage{microtype}      
\usepackage[table]{xcolor}
\usepackage[pdftex]{graphicx}
\usepackage{subcaption} 
\usepackage{wrapfig}
\usepackage{float}
\usepackage{listings}
\definecolor{lightgray}{rgb}{.9,.9,.9}
\definecolor{darkgray}{rgb}{.4,.4,.4}
\definecolor{purple}{rgb}{0.65, 0.12, 0.82}
\lstdefinelanguage{JavaScript}{
  keywords={break, case, catch, continue, debugger, default, delete, do, else, false, finally, for, function, if, in, instanceof, new, null, return, switch, this, throw, true, try, typeof, var, void, while, with},
  morecomment=[l]{//},
  morecomment=[s]{/*}{*/},
  morestring=[b]',
  morestring=[b]",
  ndkeywords={class, export, boolean, throw, implements, import, this},
  keywordstyle=\color{blue}\bfseries,
  ndkeywordstyle=\color{darkgray}\bfseries,
  identifierstyle=\color{black},
  commentstyle=\color{purple}\ttfamily,
  stringstyle=\color{red}\ttfamily,
  sensitive=true
}

\title{Cropland Mapping using Geospatial Embeddings}

\author{
Ivan Zvonkov$^{1}$ \quad Gabriel Tseng$^{2,3}$ \quad Inbal Becker-Reshef$^{1}$ \quad Hannah Kerner$^{4}$ \\
$^1$ University of Maryland, College Park \\ $^2$ Mila -- Quebec AI Institute \\ $^3$ McGill University \\ $^4$ Arizona State University 
}

\begin{document}
\maketitle

\begin{abstract}
Accurate and up-to-date land cover maps are essential for understanding land use change, a key driver of climate change. Geospatial embeddings offer a more efficient and accessible way to map landscape features, yet their use in real-world mapping applications remains underexplored. In this work, we evaluated the utility of geospatial embeddings for cropland mapping in Togo. We produced cropland maps using embeddings from Presto and AlphaEarth.  Our findings show that geospatial embeddings can simplify workflows, achieve high-accuracy cropland classification and ultimately support better assessments of land use change and its climate impacts.
\end{abstract}

\section{Introduction \& Related Work}

Human-driven land transformation is the most substantial human alteration of the Earth \cite{HumanDomination}. Researchers have used Earth observation data to better understand land use change (e.g., measuring global cropland expansion using the Landsat satellite data archive  \cite{GLAD}). The amount of data that needs to be fetched and processed presents a major bottleneck for land use and land change mapping efforts. 

Geospatial embeddings provide a fast and scalable way to turn Earth observation data into actionable maps. They are created by first training a model with self-supervised learning and then running large-scale inference over a region of interest  (e.g., \cite{TESSERA}). Models that generate geospatial embeddings are often called geospatial foundation models (GeoFMs). While many GeoFMs have been released \cite{AnySat, CROMA, DOFA, MMEarth, SatMAE, ScaleMAE, SoftCon, Prithvi-100M-preprint}, their use in real-world applications remains limited \cite{SustainFM}.

In this paper, we explored the use case of cropland mapping using geospatial embeddings. We generated cropland maps in Togo using embeddings from Presto \cite{Presto} and AlphaEarth \cite{AlphaEarth}. We used Presto because it has relatively low compute requirements while maintaining strong performance across Earth observation tasks. We used AlphaEarth \cite{AlphaEarth} embeddings because they have already been generated globally. To evaluate utility, we analyzed the quality of the generated cropland maps and compared them to existing products. By lowering barriers to generating accurate cropland maps, geospatial embeddings can support more timely monitoring of land use change and its impact on the climate.

\section{Methodology}

\subsection{Embedding generation}

We generated embeddings for all of Togo (56,785 km$^2$) for the time frame of March 2019 - March 2020. Embedding generation took 16 hours and cost \$313.40. The final Togo asset size was 128.8 GB. The Google Earth Engine (GEE) asset and all relevant scripts are available at: \hyperlink{https://nasaharvest.github.io/presto-embeddings}{https://nasaharvest.github.io/presto-embeddings}. 

The embedding generation pipeline involved two steps: model deployment and inference. For model deployment, we wrote a Google Colab notebook to package Presto into a TorchServe Docker container and deploy the container to Google Cloud's Vertex AI. For inference, we created a GEE script to obtain Earth observation data for all of Togo and make predictions from the data using the \texttt{ee.Model.fromVertexAi} function. We used the Presto encoder to compress location information, optical imagery (Sentinel-2), radar imagery (Sentinel-1), climatology data (ERA5), and elevation data (SRTM) over the course of a year. Each generated Presto embedding contained 128 features representing a single 10m pixel on Earth. 

To verify that our generated embeddings contained meaningful information, we clustered the embeddings and visually compared them to the WorldCover land cover map \cite{WorldCoverMap}, which has been shown to have accurate cropland classification in Togo \cite{HowAccurate}. 
We performed clustering using the \texttt{ee.Clusterer.wekaKMeans} algorithm available in GEE with 7 classes. We sampled 1000 points from our generated embeddings to train the clustering algorithm.

\begin{figure}[H]
    \centering
    \includegraphics[width=0.9\textwidth]{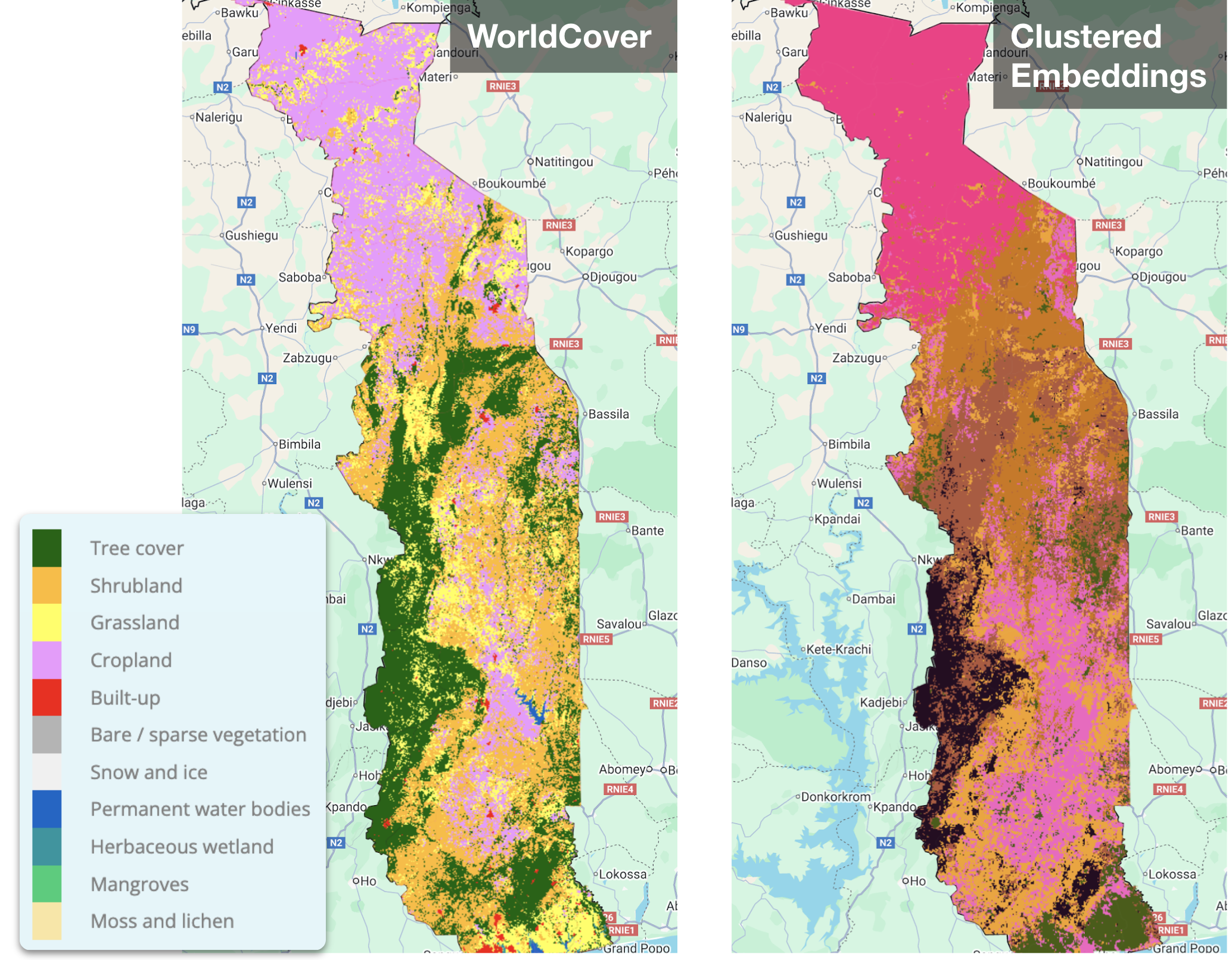} 
    \caption{Comparison of WorldCover land cover map and clustered embeddings in Togo. We used the default WorldCover color scheme. The cluster colors were randomly generated.}
    \label{fig:clusters}
\end{figure}

Figure \ref{fig:clusters} showed that the Presto embedding clusters were visually aligned with the land cover map. In particular, the green embedding cluster correlated closely with built-up areas in the land cover map. This gave us confirmation that the embeddings captured meaningful information about the land cover in Togo and could be used for further mapping and analysis.

\subsection{Mapping using embeddings}

\begin{figure}
  \centering
  \includegraphics[width=0.95\linewidth]{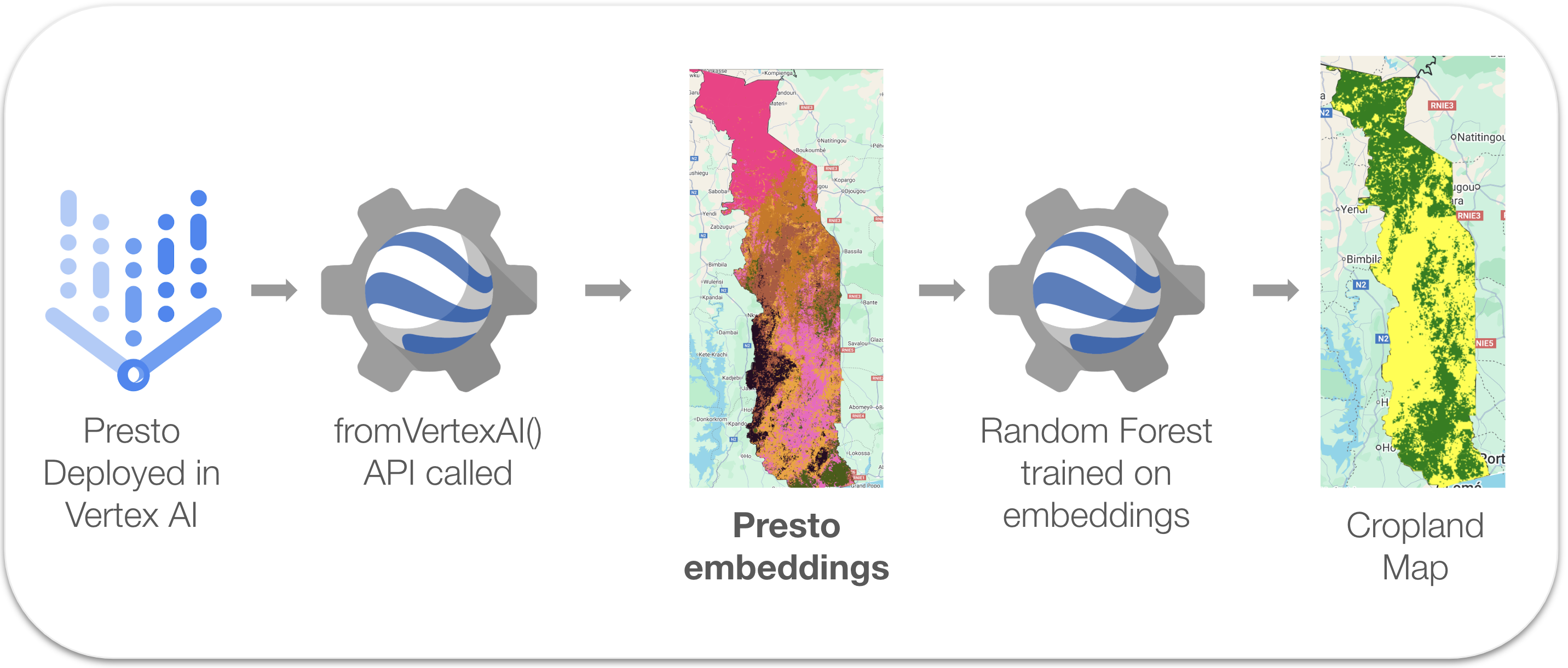}
  \caption{Steps involved in generating embeddings and mapping cropland.}
  \label{fig:pipeline}
\end{figure}

Our embedding approach allowed us to train a classifier using \textbf{8 lines of code} (see the GEE script below). When training a classifier on top of the embeddings, we used labeled points from the Togo cropland classification task in CropHarvest \cite{CropHarvest}. We evaluated the classifier's outputs using labels collected in March 2020 as part of a Rapid Response crop mapping effort in Togo \cite{TogoRapidResponse}.  We used the default GEE Random Forest classifier initialized with 100 trees. We trained the classifier on a training dataset consisting of Togo labels paired with embeddings. The initial outputs contained extensive crop predictions, so we applied a probability threshold of 0.7 to reduce false positives. The resulting cropland map was generated in less than 5 seconds. The entire script is included below:

\begin{lstlisting}[language=Javascript]
// 1. Load embeddings for region of interest
var roi = ee.FeatureCollection("FAO/GAUL/2015/level2").filter("ADM0_NAME=='Togo'");
var embeddings = ee.Image("users/<omitted for review>/Togo/Presto_embeddings_v2025_06_19")

// 2. Load Togo points
var points = ee.FeatureCollection("users/<omitted for review>/Togo/points_2019")
var trainPoints = points.filter(ee.Filter.eq("subset", "training"))

// 3. Create training dataset (training points + embeddings)
var trainSet = embeddings.sampleRegions(trainPoints, ["is_crop"], 10)

// 4. Train a classifier (100 trees)
var model = ee.Classifier.smileRandomForest(100).setOutputMode('probability').train(trainSet, 'is_crop', embeddings.bandNames());

// 5. Classify embeddings using trained model
var croplandPreds = embeddings.classify(model).clip(roi)
var croplandMap = croplandPreds.gte(0.7).rename("map_crop")
\end{lstlisting}

Our methodology is illustrated in Figure \ref{fig:pipeline}. For comparison, we applied the same cropland mapping process to the AlphaEarth embeddings \cite{AlphaEarth} (comparison of Presto and AlphaEarth embeddings is shown in Table \ref{tab:embeddings_comparison}).

\begin{table}[H]
\centering
\begin{tabular}{p{3cm} p{4.5cm} p{4.5cm}}
\toprule
 & \textbf{Presto Embeddings} & \textbf{AlphaEarth  Embeddings} \\
\midrule
\textbf{Data Sources} & Sentinel-1, Sentinel-2, ERA5, SRTM & Optical, radar, LiDAR, and other sources \\
\textbf{Time frame} & March 2019 -- March 2020 & January 2021 -- January 2022 \newline [earlier than 2021 not available] \\
\textbf{Embedding Size} & 128 values & 64 values \\
\textbf{Scale} & 10 m & 10m \\
\textbf{Data Type} & \verb|uint16| & \verb|double| \\
\textbf{Bytes per pixel} & 256 & 512 \\
\bottomrule
\end{tabular}
\vspace{4pt}
\caption{Comparison of Presto and AlphaEarth embeddings.}
\label{tab:embeddings_comparison}
\end{table}


\section{Results \& Conclusion}

\begin{figure}
  \centering
  \begin{minipage}{0.2\textwidth}
    \includegraphics[width=\linewidth]{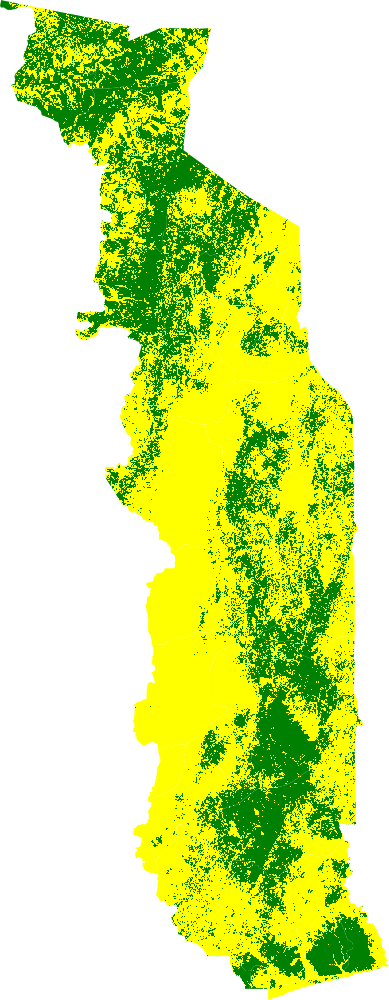}
    \caption*{\hspace{2.5em}GLAD \cite{GLAD}}
    \label{fig:second}
  \end{minipage}
  \hfill
  \begin{minipage}{0.2\textwidth}
    \includegraphics[width=\linewidth]{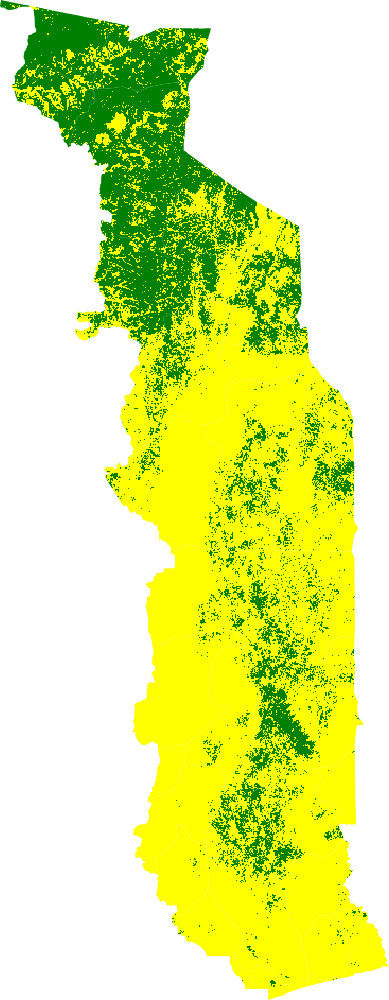}
    \caption*{\hspace{1.2em}WorldCover \cite{WorldCoverMap}}
    \label{fig:first}
  \end{minipage}
  \hfill
  \begin{minipage}{0.2\textwidth}
    \includegraphics[width=\linewidth]{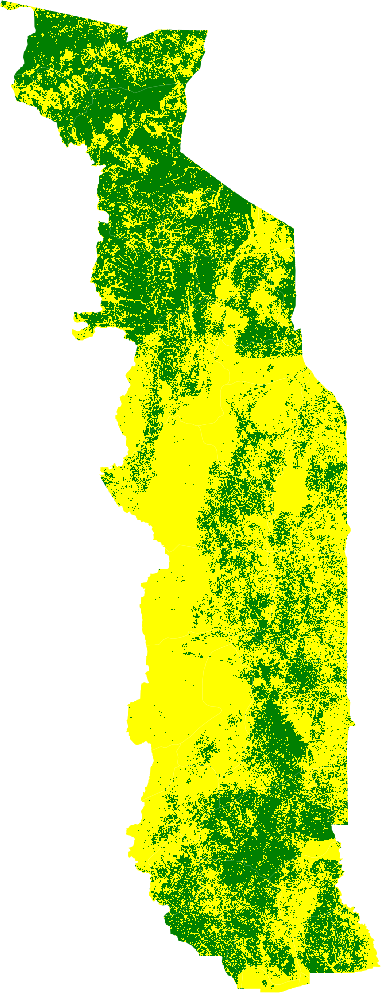}
    \caption*{\hspace{2.5em}\makecell{Presto\\ Embeddings}}
    \label{fig:third}
  \end{minipage}
  \hfill
  \begin{minipage}{0.2\textwidth}
    \includegraphics[width=\linewidth]{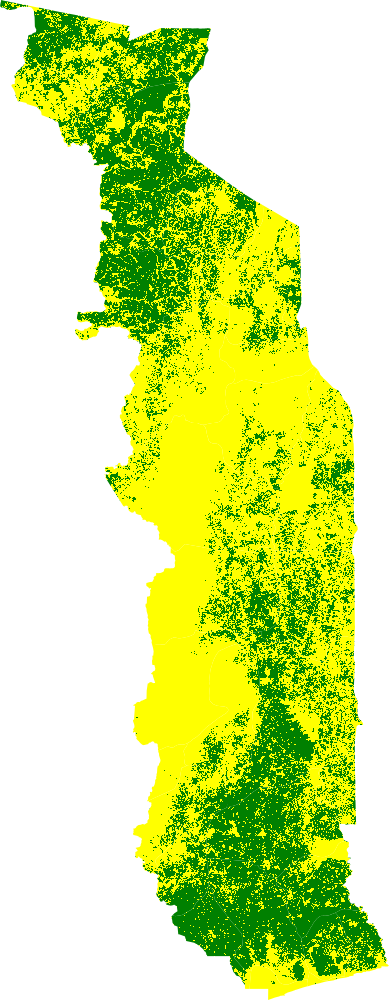}
    \caption*{\hspace{2.5em}\makecell{AlphaEarth\\ Embeddings}}
    \label{fig:fourth}
  \end{minipage}
  \caption{Comparison of cropland maps in Togo. Green is cropland and yellow is not cropland.}
  \label{fig:maps}
\end{figure}

We conducted a visual assessment on the generated cropland maps and compared them to two existing cropland maps (Figure \ref{fig:maps}). For the existing cropland maps, we chose GLAD \cite{GLAD} and WorldCover \cite{WorldCoverMap} because they have high cropland accuracy in Togo \cite{HowAccurate}. We found that Presto and AlphaEarth embedding-based maps have similar cropland proportion and distribution. We also noted that Presto and AlphaEarth embedding-based maps classify more pixels as cropland than WorldCover. Appendix \ref{app:analysis} provides additional qualitative analysis.

To get a more comprehensive understanding of the map quality, we conducted an accuracy assessment using the Togo test set from CropHarvest \cite{CropHarvest}. It's important to note that both Presto and AlphaEarth embeddings were benchmarked against data in the Togo test set. The Presto embeddings-based cropland map achieved the highest overall accuracy, producer's accuracy, and F1 score (Table \ref{tab:cropland-metrics}).

\begin{table}[H]
  \begin{tabular}{
    >{\raggedright\arraybackslash}p{2.9cm}  
    >{\centering\arraybackslash}p{2.2cm}  
    >{\centering\arraybackslash}p{2.2cm}
    >{\centering\arraybackslash}p{2.2cm}
    >{\centering\arraybackslash}p{2.2cm}
  }
    \toprule
    & GLAD \cite{GLAD}
    & WorldCover \cite{WorldCoverMap} 
    & \makecell[c]{Presto\\embeddings} 
    & \makecell[c]{AlphaEarth\\embeddings} \\
    \midrule
    Overall Accuracy    & 0.859 & 0.880 & \textbf{\textcolor{blue}{0.897}} & 0.859 \\
    User's Accuracy     & 0.821 & \textbf{\textcolor{blue}{0.892}} & 0.833 & 0.745 \\
    Producer's Accuracy & 0.627 & 0.647 & \textbf{\textcolor{blue}{0.784}} & 0.745 \\
    F1 Score            & 0.711 & 0.750 & \textbf{\textcolor{blue}{0.808}} & 0.745 \\
    \bottomrule
  \end{tabular}
  \centering
  \vspace{4pt}
    \caption{Cropland metrics for Togo.}
    \label{tab:cropland-metrics}
\end{table}

\paragraph{Conclusion}
We showed that using embeddings for cropland mapping is simple, quick, and effective. While our experiments are limited to Togo, they serve as a case study to illustrate the methodology. Future work will extend this evaluation across additional regions and applications. Our results highlight the potential for using embeddings for more effective mapping and demonstrate the growing potential of embeddings for understanding land use change and its impact on the climate.




\bibliography{references}

\appendix

\section{Appendix}

\subsection{Additional Qualitative Analysis} \label{app:analysis}
\begin{figure}
    \centering
  \includegraphics[width=0.7\textwidth]{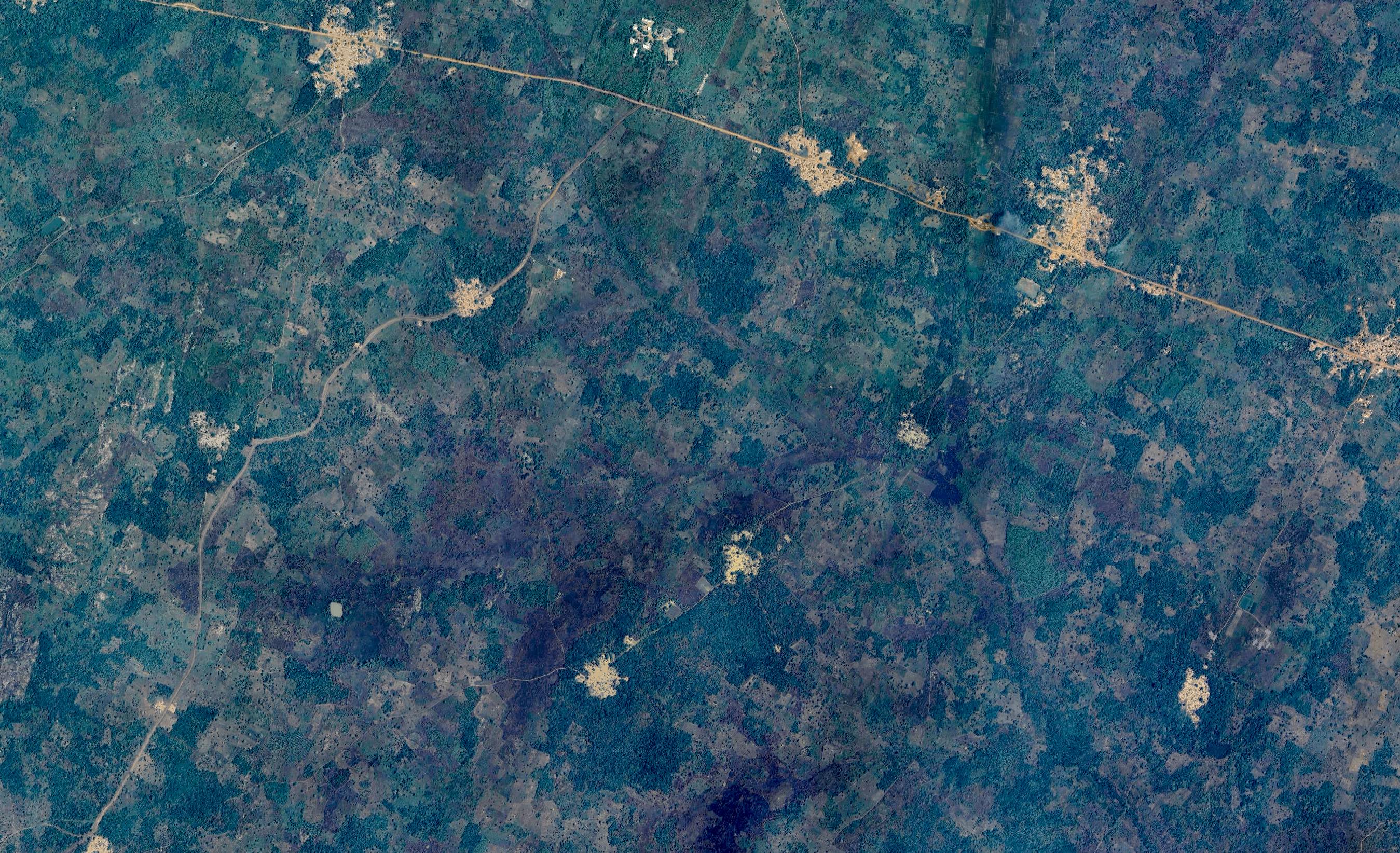} 
    \caption{High resolution satellite image zoomed in on coordinate: (0.84445, 6.46924).}
    \label{fig:Togo_zoom_satellite1}
\end{figure}

\begin{figure}
    \centering
    \includegraphics[width=0.7\textwidth]{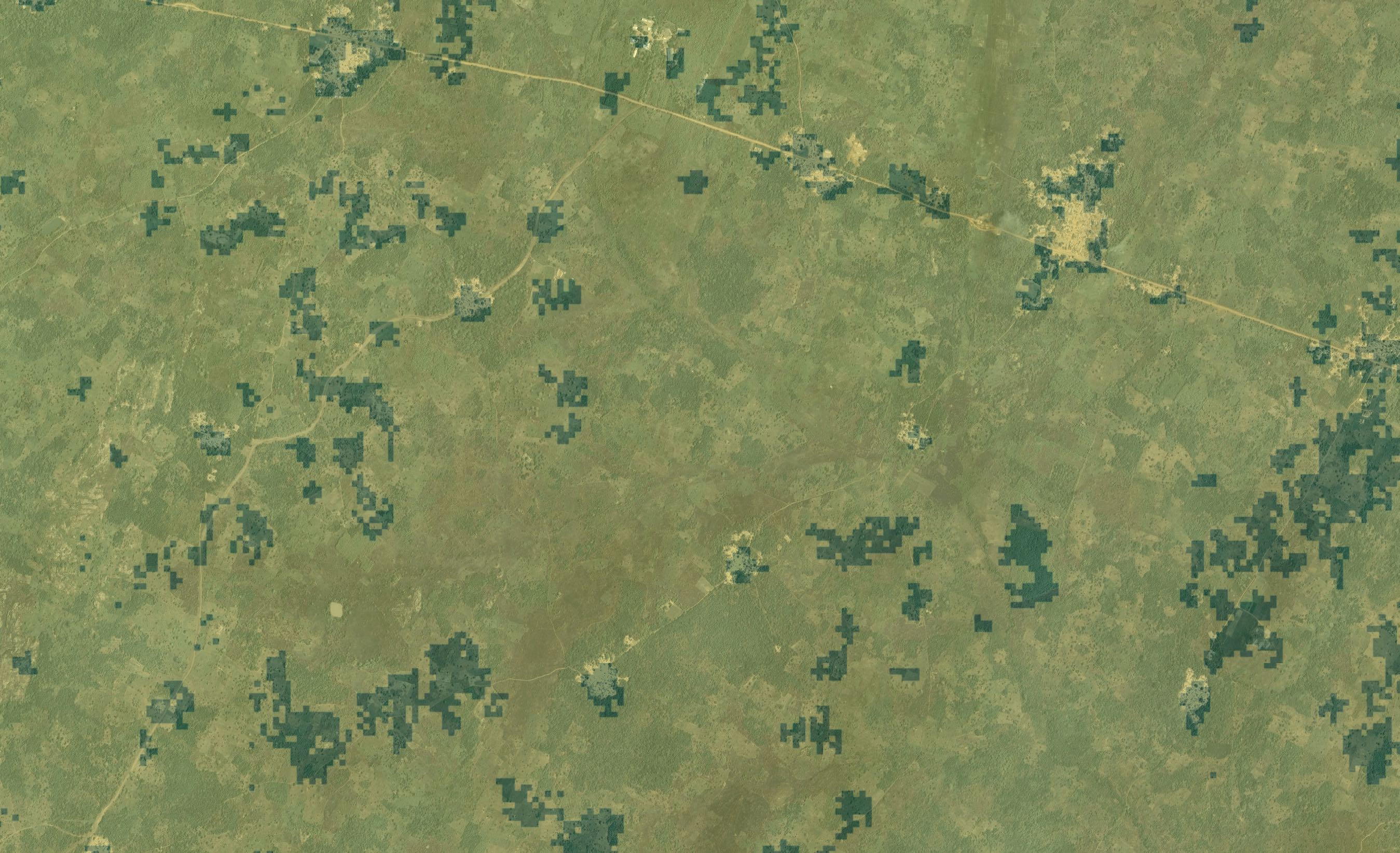} 
    \caption{GLAD cropland map \cite{GLAD} zoomed in on coordinate: (0.84445, 6.46924).}
    \label{fig:Togo_zoom_GLAD}
\end{figure}

\begin{figure}
    \centering
  \includegraphics[width=0.7\textwidth]{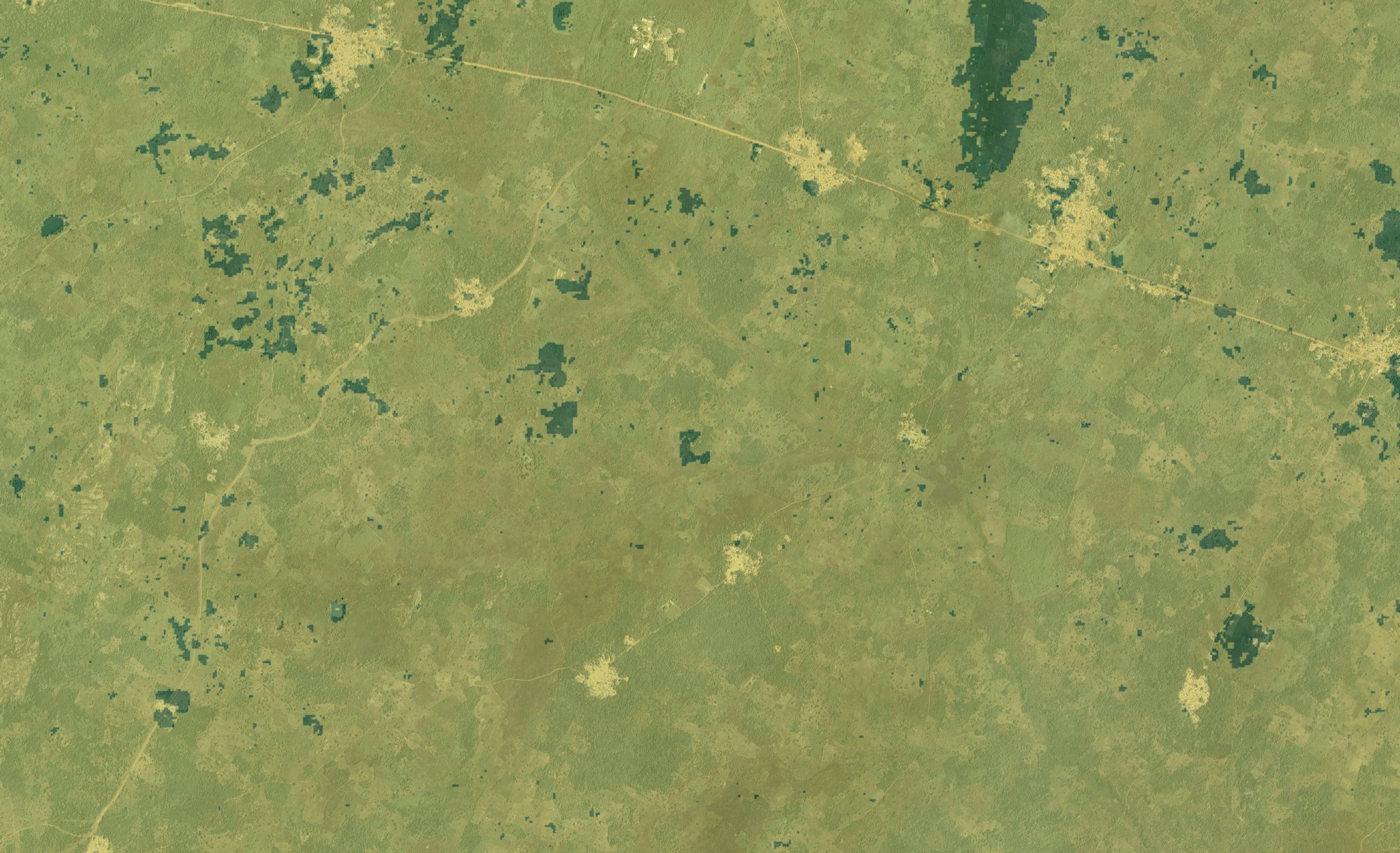} 
    \caption{WorldCover cropland map \cite{WorldCoverMap} zoomed in on coordinate: (0.84445, 6.46924).}
    \label{fig:Togo_zoom_WorldCover}
\end{figure}

\begin{figure}
    \centering
  \includegraphics[width=0.7\textwidth]{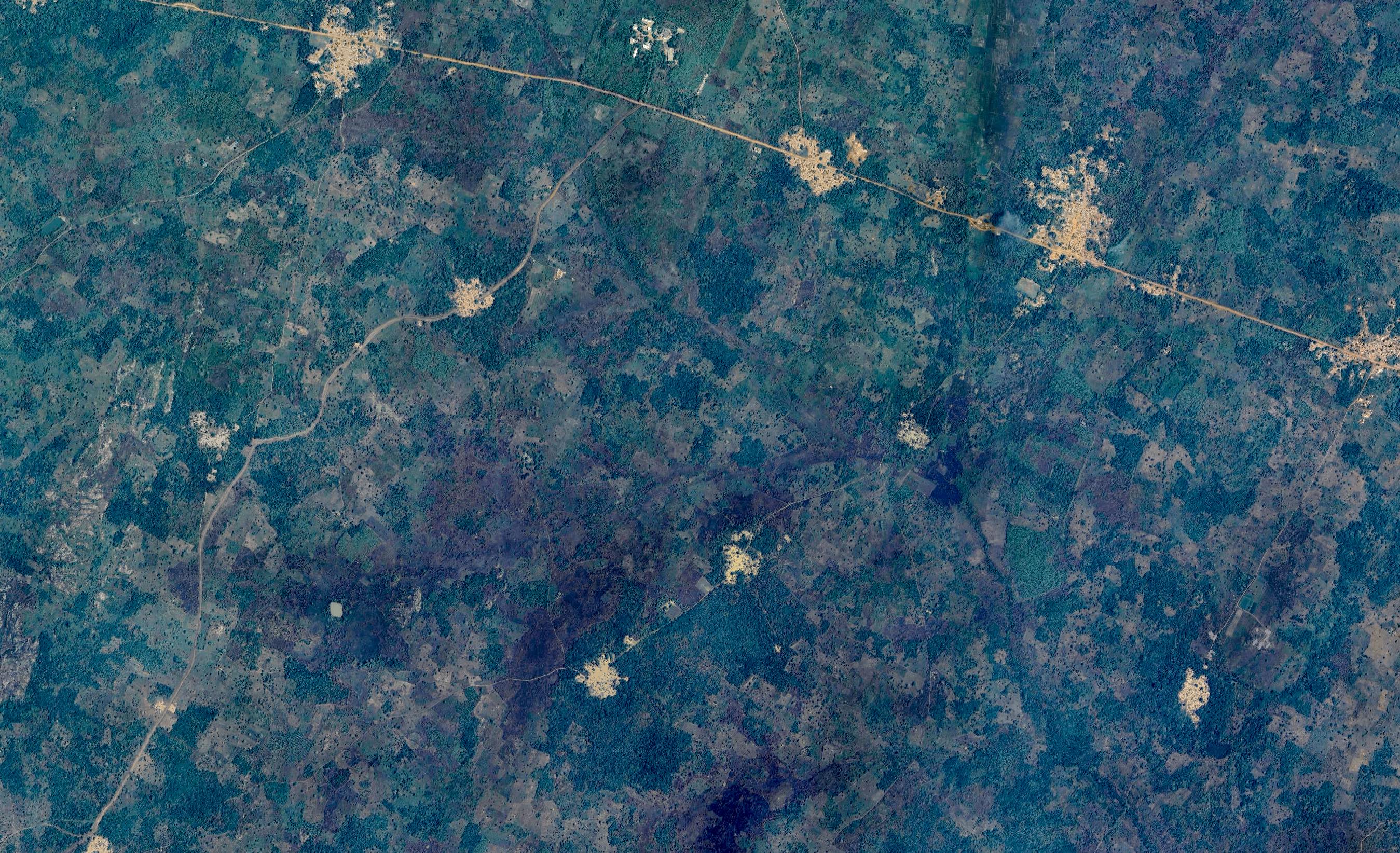} 
    \caption{High resolution satellite image zoomed in on coordinate: (0.84445, 6.46924).}
    \label{fig:Togo_zoom_satellite2}
\end{figure}

\begin{figure}
    \centering
  \includegraphics[width=0.7\textwidth]{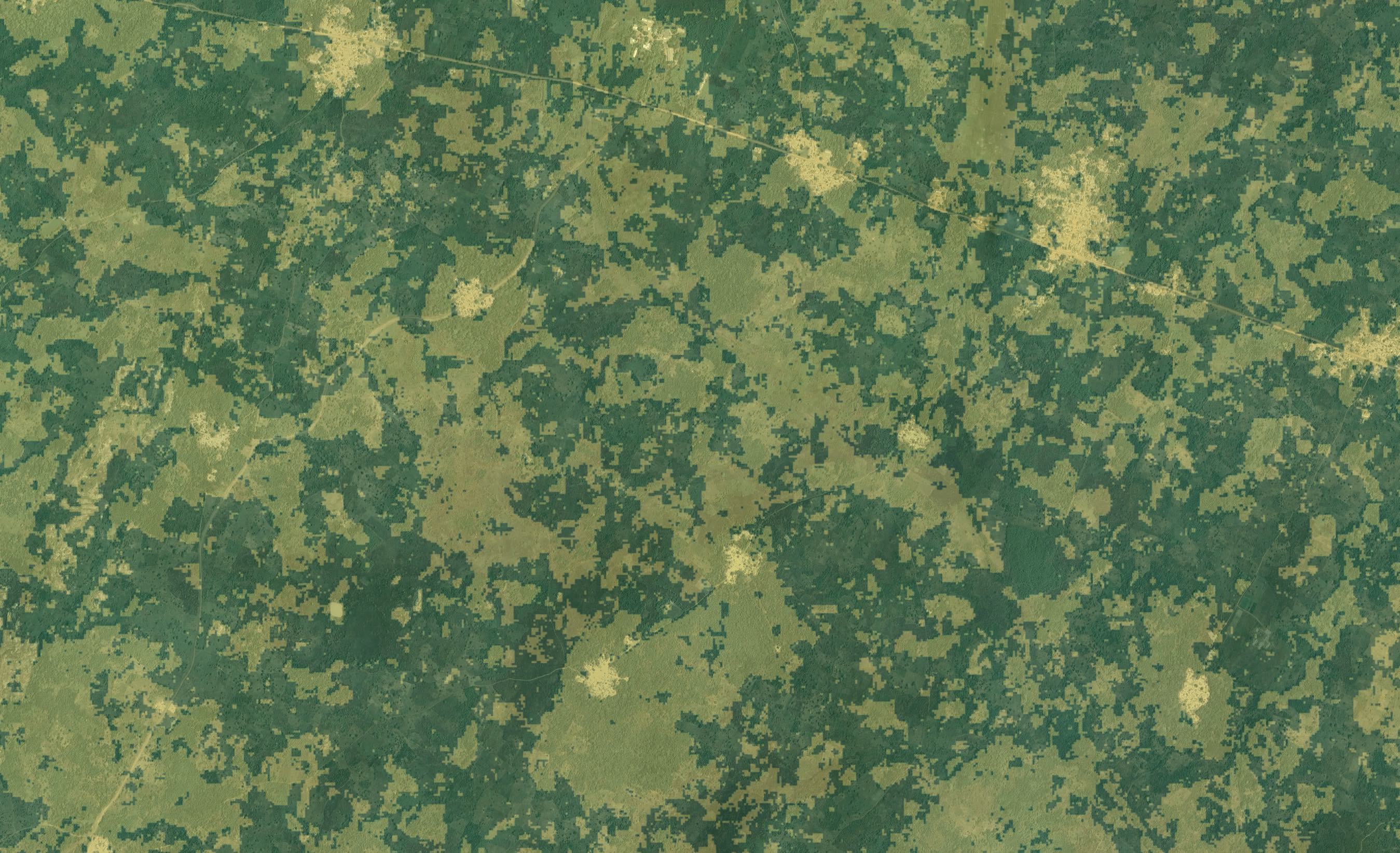} 
    \caption{Presto cropland map zoomed in on coordinate: (0.84445, 6.46924).}
    \label{fig:Togo_zoom_Presto}
\end{figure}

\begin{figure}
    \centering
    \includegraphics[width=0.7\textwidth]{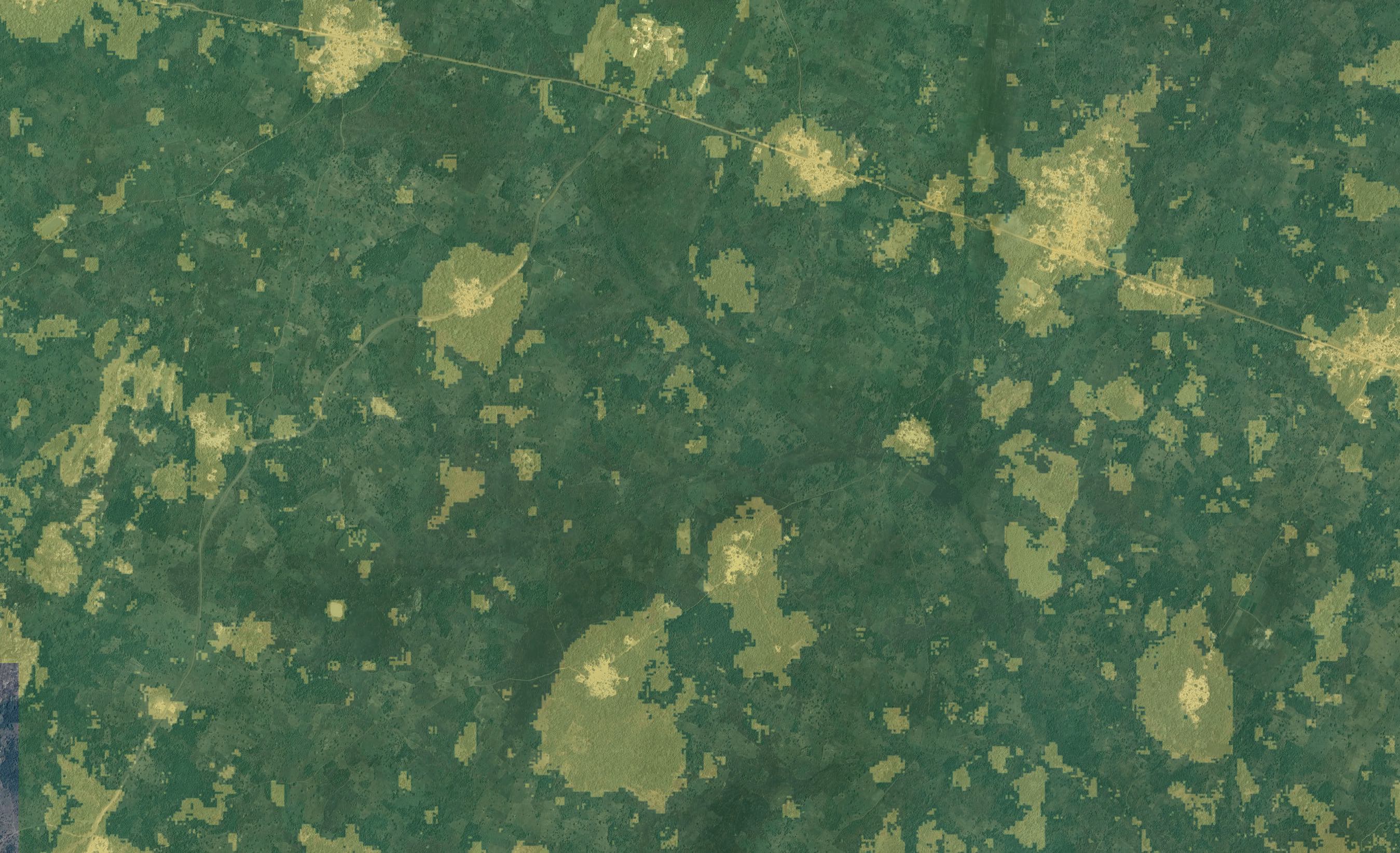} 
    \caption{AlphaEarth cropland map zoomed in on coordinate: (0.84445, 6.46924).}
    \label{fig:Togo_zoom_DeepMind}
\end{figure}

We conducted further investigation into the South of Togo where all maps have differences. We zoomed into a window centered on coordinate: (0.84445, 6.46924) and compared a high resolution satellite image (Figures \ref{fig:Togo_zoom_satellite1}, \ref{fig:Togo_zoom_satellite2}) to all the cropland maps (Figures \ref{fig:Togo_zoom_WorldCover}, {\ref{fig:Togo_zoom_GLAD}}, \ref{fig:Togo_zoom_Presto}, \ref{fig:Togo_zoom_DeepMind}).
In this window we found that all maps with the exception of GLAD correctly identify built-up areas as non-crop. Embedding-based maps appear to classify more cropland (green) correctly. AlphaEarth embedding-based map over predicts crops in some shrub/tree areas, while the Presto embedding-based map visually accurately identifies many of the tree areas as non-crop.
\clearpage

\end{document}